\documentclass[review]{elsarticle}

\usepackage{lineno,hyperref}
\modulolinenumbers[5]

\usepackage{graphicx}
\usepackage{threeparttable}

\RequirePackage{amsthm,amsmath}
\RequirePackage[numbers]{natbib}
\usepackage{bm}

\usepackage[printwatermark]{xwatermark}
\usepackage{xcolor}
\usepackage{graphicx}
\usepackage{lipsum}

\journal{Computational Statistics and Data Analysis}

\begin{document}

\begin{frontmatter}

\title{Display advertising: Estimating conversion probability efficiently}
%\tnotetext[mytitlenote]{Fully documented templates are available in the elsarticle package on \href{http://www.ctan.org/tex-archive/macros/latex/contrib/elsarticle}{CTAN}.}

%% Group authors per affiliation:
\author[mymainaddress]{Abdollah Safari\corref{mycorrespondingauthor}}
%\address{Radarweg 29, Amsterdam}
%\fntext[myfootnote]{Since 1880.}

%\author[mysecondaryaddress]{Abdollah Safari\corref{mycorrespondingauthor}}
\cortext[mycorrespondingauthor]{Corresponding author}
\ead{asafari@sfu.ca}

%% or include affiliations in footnotes:
\author[mymainaddress]{Rachel MacKay Altman}
\author[mymainaddress]{and Thomas M. Loughin}
%\ead[url]{www.elsevier.com}

\address[mymainaddress]{Department of Statistics and Actuarial Science\\
Simon Fraser University\\
8888 University Drive\\
Burnaby, BC V5A 1S6\\
CANADA}
%\address[mysecondaryaddress]{360 Park Avenue South, New York}

\begin{abstract}
The goal of online display advertising is to entice users to ``convert" (i.e., take a pre-defined action such as making a purchase) after clicking on the ad. An important measure of the value of an ad is the probability of conversion. The focus of this paper is the development of a computationally efficient, accurate, and precise estimator of conversion probability. The challenges associated with this estimation problem are the delays in observing conversions and the size of the data set (both number of observations and number of predictors). Two models have previously been considered as a basis for estimation: A logistic regression model and a joint model for observed conversion statuses and delay times. Fitting the former is simple, but ignoring the delays in conversion leads to an under-estimate of conversion probability. On the other hand, the latter is less biased but computationally expensive to fit. Our proposed estimator is a compromise between these two estimators. We apply our results to a data set from Criteo, a commerce marketing company that personalizes online display advertisements for users.
\end{abstract}


\end{frontmatter}

%\linenumbers

\section{Introduction}

Display advertising is a relatively new type of online advertisement where advertisers pay publishers to present their ads (also known as impressions) on different webpages. Depending on the purpose of the advertisement, different payment options can be used. These options include cost per impression, where the advertisers pay the publishers to display their ads (whether the user clicks the ad or not), cost per click, where the advertisers pays for an impression only if a user clicks on it, and cost per action (CPA), where advertisers pay only if the user takes a predefined action (conversion) after clicking the ad, such as purchasing a product or service \cite{Muthuk, Chapelle}.

For profitability, the CPA option requires that publishers make a ``good" match between advertisers and customers. In particular, they should display ads with high expected earnings per impression, i.e., ads where the customer's probability of clicking and the subsequent probability of the click's converting are high. \cite{McAfee}. The entire process of ad selection needs to be completed in the time between when a user opens a page and when the page is fully rendered. Thus, the publisher has a very short time in which to choose which ad(s) to display to the user. Great progress has been made predicting whether a user will click on an impression in the context of search advertising (see for example \citet{Hillard10}, or \citet{McMahan12}) and display advertising (see for example \citet{Chapelle14}, or \citet{Agarwal10}). However, little is known about estimating the probability of conversion. For instance, \cite{Rosales12} perform an experimental analysis (on a private Yahoo data set) to show the advantage of conversion probability over the click probability as a measure of profitability in display advertising, and point out the lack of inference about this new measurement in the literature.

The main issue in conversion probability estimation is the delay between the click and the eventual conversion status of the click (called the conversion delay), which can vary from a few milliseconds to months. In other words, eventual conversion status (converted or unconverted) is unknown for clicks where the conversion delay is censored. \citet{Chapelle} proposed using the maximum likelihood estimator (MLE) of the conversion probability based on a delay feedback model (DFM), a mixture model for observed conversion status that depends on the delay distribution. Although his estimator is accurate when the model is correctly specified, his approach is not computationally efficient. Efficiency is critical in this Big Data setting where publishers need to re-estimate conversion probability rapidly and frequently as time progresses and more data accrue (i.e., real-time updating). In addition, the performance of his estimator is unknown when the delay distribution is not exponential.

Our goal in this paper is to develop a method for estimating probability of conversion with high accuracy and in a computationally efficient manner. In particular, we introduce a new estimator based on the logistic regression model that (wrongly) treats all conversion statuses as known, and then reduce the bias of this estimator through a novel application of the Kullback-Leibler distance. We evaluate the accuracy and computational efficiency of this new estimator compared to those of Chapelle's estimator. In addition, we study the performance of these estimators when the delay distribution is misspecified.

In \S\ref{Model} we define some notation and present the DFM of \citet{Chapelle}. In \S\ref{Est}, we introduce our estimator along with an algorithm to evaluate it efficiently for a given data set. Section \S\ref{application} presents an application of our results to a data set released by Criteo \cite{Chapelle}, and \S\ref{Simulation} describes a simulation study that illustrates the accuracy, precision, and computational efficiency of the estimators.

\section{Model specification}
\label{Model}
In this section, we describe the DFM developed by \citet{Chapelle}. The assumptions of the DFM (and of our methods that follow) are: \textbf{i.} the true conversion probability is fixed over time, \textbf{ii.} the predictors don't depend on time, \textbf{iii.} a converted click can never become unconverted, \textbf{iv.} an unconverted click with delay time less than a fixed time period (the conversion window) can convert, \textbf{v.} an unconverted click with delay time greater than the conversion window cannot convert (in other words, an unconverted click can convert any time within the conversion window, $W$), and \textbf{vi.} $W$ is long enough that only a negligible proportion of conversions occur outside this window.

Let the data collection start at time $0$. Label clicks sequentially in time as $1, 2, 3, \dots$. Let $t_{i,0}$ be the time of click $i$ -- treated as non-random for the purposes of this paper. Throughout, we use bold letters to denote vectors. For instance, $\bm{x_i}\equiv (x_{i,1},\ldots,x_{i,k})$ is a $1 \times k$ vector of covariates associated with click $i$, e.g., attributes of the user and/or origin website. We define $x_{i,1} = 1~\forall i$ so as to include an intercept. Define $C_i$ to be the eventual conversion status indicator for click $i$, i.e. $C_i = 1$ if click $i$ ever converts and $C_i = 0$ otherwise. Let $T_i^c$ be the time at conversion if $C_i = 1$; if $C_i = 0$, then fix $T_i^c = t_{i,0} + W$. Then the delay time $D_i$ is defined as $D_i = T_i^c - t_{i,0}$ (so that $D_i = W$ if $C_i = 0$). Given $\bm{x_i}$ and $C_i = 1$, let $h_i(d) = h(d~|~\bm{x_i}, C_i=1)$ and $H_i(d) = H(d~|~\bm{x_i}, C_i=1)$ be the conditional pdf and cdf, respectively, of $D_i$.

%Assume that the click is still within the conversion window. 

Now suppose that at a given moment $t > 0$ we wish to estimate the conversion probability of a click with covariates $\bm{x_i}$. Define $a_i(t) = \min\{t - t_{i,0} , W\}$ to be the age of click $i$. Since we treat $t_{i,0}$ as non-random, $a_i(t)$ is non-random as well.

For subsequent derivations, we consider a given fixed time $t$ and suppress $t$ in our notation for convenience. At this time, say $n$ clicks have accumulated. Let $Y_i$ be the current conversion status indicator of click $i = 1, \dots, n$, i.e., $Y_i = 1$ if click $i$ converted prior to time $t$ and $Y_i = 0$ otherwise. Note that $D_i$ is observed prior to $t$ if $Y_i = 1$, and is greater than or equal to $a_i$ (right censored) if $Y_i = 0$.

To the best of our knowledge, the DFM is the only model for conversion probability in the literature that incorporates conversion delays, i.e., that is based on the bivariate response for each click, 
$(Y_i,D_i)$. In this model, $C_i$ is assumed to follow a logistic regression model with $p_i=P(C_i = 1 | \bm{x_i})= \frac{exp(\bm{\beta_c} ^ \prime \bm{x_i})}{1 + exp(\bm{\beta_c} ^ \prime \bm{x_i})}$. Given $C_i = 1$ and $\bm{x_i}$, delay times are assumed to follow an exponential distribution with rate $\lambda_i= \exp(\bm{\beta_c} ^ \prime \bm{x_i})$. The log-likelihood function of the DFM is then
\begin{eqnarray*}
\l\left( \bm{\beta_c} , \bm{\beta_d} | \bm{y} , \bm{d} \right) = &-& \sum_{i: y_i = 1} \{ \log\left[ P(C_i = 1 | \bm{x_i})\right] + \log(\lambda_i) - \lambda_i d_i \} \\ \nonumber
&-& \sum_{i: y_i = 0} \log\left[1 - P(C_i = 1 | \bm{x_i}) + P(C_i = 1 | \bm{x_i}) \exp(-\lambda_i a_i) \right] \nonumber
\end{eqnarray*}

For later derivations in this paper, we will require a different form for $\l$. Specifically, we define $Z_i$ as
\begin{eqnarray}
Z_{i}(a_{i}) \equiv Z_{i} := \min (D_{i} , a_{i}),
\end{eqnarray}
so that $0 \leq Z_{i} \leq a_{i}$. Note that $Z_i$ is a function of a single random variable, $D_{i}$. We can define an equivalence relationship between $Y_{i}$ and $Z_i$ as
\begin{eqnarray}
Z_{i}  <  a_{i} \iff Y_{i} = 1 \\ \nonumber
Z_{i}  = a_{i} \iff Y_{i} = 0
\end{eqnarray}
Then, the likelihood function of the DFM can be rewritten in terms of the $z_i$'s (realizations of the $Z_i$'s) as
\begin{eqnarray}
\label{TrueModel}
L_g \left(\bm{\beta_c} | \bm{z} \right) = \prod_{i} \left( p_i h(z_{i}) \right) ^ {I( z_{i} < a_{i})} \left( 1 - p_i H(z_{i}) \right) ^ {I( z_{i} \geq a_{i})}.
\end{eqnarray}
(See Appendix \ref{appendix} for the proof.)

\section{Estimation}
\label{Est}

As discussed by \citet{Chapelle}, the likelihood function in \eqref{TrueModel} is non-convex with no closed form for the MLE. Therefore, its optimization is very slow. For this reason, we consider alternative estimators in this section.

\subsection{Naive estimator}

A simple (but misspecified model) for observed conversion status is the logistic regression model where the current conversion statuses of the clicks are treated as their eventual conversion statuses. In other words, conversion delay time (and the possibility that unconverted clicks with age less than $W$ could convert) are ignored. \citet{Chapelle} calls this model the ``naive model". The likelihood function of this model is
\begin{eqnarray}
\label{MissModel}
L_f \left(\bm{\alpha} | \bm{z} \right) &=& f \left( \bm{z} | \bm{\alpha} \right) \\ \nonumber
&=& \prod_{i = 1}^{n} f \left( z_{i} | \bm{\alpha} \right) \\ \nonumber
&=& \prod_{i = 1}^{n} \left[ \theta_i ^ {I( z_{i} < a_{i})} (1 - \theta_i) ^ {I( z_{i} \geq a_{i})} \right] ,
\end{eqnarray}
where $\theta_i = \frac{\exp(\bm{\alpha}^\prime \bm{x_i})}{1 + \exp(\bm{\alpha}^\prime \bm{x_i})}$ is the conversion probability of the $i^{th}$ click and $\bm{\alpha}$ is a vector of regression coefficients. The likelihood function of the naive model is convex and computationally efficient to optimize. However, the MLE of $\theta_i$ is biased low for the true probability of conversion, since some unconverted clicks could convert later.

\subsection{Bias - adjusted estimator} \label{BAestimator}

In this section, we introduce a new estimator to adjust for the bias in the naive estimator. We use the Kullback-Leibler information criterion (KLIC) approach \cite{White82}. Suppose that $g(\bm{z} | \bm{\beta_c})$ is the true data-generating distribution, but that $f(\bm{z} | \bm{\alpha})$ is the assumed model. Then the KLIC can be computed as follows:
\begin{eqnarray*}
KLIC \left( g : f ~ ; ~ \bm{\alpha}, \bm{\beta_c} \right) &=& E_g \left( \ln \left[ \frac{g(\bm{z} | \bm{\beta_c})}{f(\bm{z} | \bm{\alpha})} \right] \right) \\ \nonumber
&=& E_g \left( \ln \left[g(\bm{z} | \bm{\beta_c}) \right] \right) - E_g \left( \ln \left[f(\bm{z} | \bm{\alpha}) \right] \right) \\ \nonumber
&=& E_g \left( \ln \left[g(\bm{z} | \bm{\beta_c}) \right] \right) - E_g \left( \ln \left[ \prod_i f(z_{i} | \bm{\alpha}) \right] \right) \\ \nonumber
&=& E_g \left( \ln \left[g(\bm{z} | \bm{\beta_c}) \right] \right) - \sum_i E_g \left( \ln \left[ f(z_{i} | \bm{\alpha}) \right] \right) \\ \nonumber
&=& E_g \left( \ln \left[g(\bm{z} | \bm{\beta_c}) \right] \right)  \\ \nonumber
&-& \sum_i \left[ p_i \ln \left( \theta_i \right) H_i(a_{i}) + \ln \left(1 - \theta_i \right) \left( 1 - H_i \left( a_{i} \right) p_i \right) \right]\nonumber
\end{eqnarray*}

\citet{White82} shows that the MLE of the parameters in the misspecified model is consistent for the minimizer of the KLIC. We use his results to adjust the naive estimator and remove its asymptotic bias relative to the true model. In other words, we assume that the true model is \eqref{TrueModel} and treat the parameters of the true model, $\bm{\beta_c}$, as known.  Then we minimize the KLIC with respect to the parameters of the misspecified model, $\bm{\alpha}$, resulting in estimating equations that depend on both $\bm{\beta_c}$ and the unknown KLIC minimizer, $\widetilde{\bm{\alpha}}$.  We then solve for $\bm{\beta_c}$.  

The details are as follows. First, we have
\begin{eqnarray}
\label{fakescore}
&& \left. \frac{\partial ~ KLIC \left( g | f ~ ; ~ \bm{\alpha} \right)}{\partial \alpha_j} \right|_{\widetilde{\bm{\alpha}}} = 0  \\ \label{MultAdj}
&\Rightarrow& \sum_i p_i H_i(a_{i}) x_{i,j} = \sum_i x_{i,j} \widetilde{\theta_i}~,~~ j = 1, \dots, k 
\end{eqnarray}
where $\widetilde{\theta_i} = \left. \theta_i \right|_{\widetilde{\bm{\alpha}}}$, and $k$ is the number of regression coefficients. Treat $H_i$ and $\widetilde{\bm{\alpha}}$ as known for the moment. Note that the equations in \eqref{MultAdj} are algebraically equivalent to the weighted quasi-score equations associated with a logistic regression model (with $\widetilde{\theta_i}$ taking the place of the usual response variable). Thus, they can be solved efficiently for $\bm{\beta_c}$. 

In the usual case where $H_i$ and $\widetilde{\bm{\alpha}}$ are unknown, we plug in consistent estimates. In particular, we compute $\hat{\theta}_i$, the MLE of $\theta_i$ from \eqref{MissModel}, which is consistent for $\widetilde{\theta_i}$ (by White's theorem).

 To estimate $H_i$, given the family of distributions of the delay (e.g., exponential), we can find the MLE of the delay distribution parameters. However, since the censoring rate could be very high, especially when $t$ is small, this MLE can be quite biased (see, e.g., \citet{Shen15}, \citet{White15}, and \citet{Hirose1999}). As a remedy, we can adjust for the delay rate estimator bias as well. \citet{Firth93} proposes a general approach to bias reduction using on a penalized score function. \citet{Pettitt98} apply Firth's approach to obtain the penalized likelihood when the responses are exponentially distributed and possibly censored.  In our notation, this penalized likelihood is 
\begin{eqnarray}
\label{ExpAdj}
L^*(\bm{\lambda} | \bm{z}) = \prod_{i \in S^*} (\lambda_i) ^ {-2} h_i(z_i) H_i(a_i),
\end{eqnarray}
where $\lambda_i = \exp(\bm{\beta_d} ^ \prime \bm{x_i})$ as before,
\begin{eqnarray}
\label{cdfexp}
%\left.
h_i(z_{i}) &=& \left\{
	\begin{array}{ll}
		\lambda_i \exp(-\lambda_i z_i),~~z_{i} < a_i \\
		\exp(-\lambda_i z_i),~~~~~z_{i} = a_i
	\end{array}
\right. ,
%\right,
\end{eqnarray}
$H_i(a_i) = 1 - \exp(-a_i \lambda_i)$, and $S^* = \{i :~C_i=1\}$. Note that since in the application, we don't know the eventual conversion status of clicks (especially for recent clicks), we approximate $S^*$ by $\Hat{S^*} = \{i :~ y_i = 1\} \cup \{i :~ y_i = 0, a_i < W\}$, which the approximation improves as time goes on. In other words, we exclude only unconverted clicks with $a_i$ longer than $W$ in \eqref{ExpAdj} since we assume they never convert, and thus don't contribute information about the delay distribution.

When the delays follow a Weibull distribution, we cannot obtain a closed form for the \citet{Firth93} penalized likelihood function. However, if we make the usual assumption that only the scale parameter of the Weibull distribution  depends on the covariates, we can obtain the Weibull penalized likelihood function for a fixed shape parameter as
\begin{eqnarray}
\label{WeibAdj}
L_w ^*(\bm{\gamma} , \nu | \bm{z}) = \prod_{i \in S^*} \left( \frac{\nu}{\gamma_i} \right) ^ 2 h^w_i(z_i) H^w_i(z_i) ,
\end{eqnarray}
where $h^w_i$ and $H^w_i$ are the pdf and cdf, respectively, of the Weibull distribution with scale parameter $\gamma_i = \exp(\bm{\beta_d} ^ t \bm{x_i})$ and shape parameter $\nu$. We suggest first estimating the shape parameter, $\nu$, by its MLE, and then treating it as a known parameter in \eqref{WeibAdj}.

We call the convergence probability estimator based on exponential and Weibull distributions for the delays the E-bias-adjusted and W-bias-adjusted estimators, respectively.

To summarize, we obtain our bias-adjusted estimate of $\bm{\beta_c}$ as follows:
\begin{enumerate}
\item Compute the MLE of $\bm{\alpha}$ based on the naive model \eqref{MissModel}.
\item Compute the maximum penalized likelihood estimates of the delay distribution parameters using Firth's approach (i.e. \eqref{ExpAdj} if the delay distribution is exponential or \eqref{WeibAdj} if the delay distribution is Weibull).
\item Compute the bias-adjusted estimate $\hat{\bm{\beta_c}}$ by solving the equations in \eqref{MultAdj}, substituting the estimates of $\bm{\alpha}$ and the delay distribution for their true values.
\end{enumerate}

Standard GLM software can be used to compute the estimates in Steps~1 and~3, while packages such as brglm in R can be used to compute the estimates in Step~2. Thus, an advantage of the bias-adjusted estimator is that it can be computed  efficiently and easily.

The similarity between \eqref{MultAdj} and the weighted quasi-score equations associated with a logistic regression model suggests that  a SE for $\hat{\bm{\beta_c}}$ (or $\hat{p}_i$, the estimator of $p_i$) could be efficiently computed as a function of the derivative of the left side of \eqref{fakescore}. We explore the validity of this SE in \S\ref{Simulation}.

\section{Application}
\label{application}

In this section, we apply our results from the previous sections to a publicly available data set released by Criteo, a commerce marketing company that connects publishers and advertisers\footnote{The data set is available at \href{http://research.criteo.com/outreach/}{http://research.criteo.com/outreach/}}. The data concern a collection of clicks that accrued over a period of two months, with $W = 30$ days \cite{Chapelle}. The eventual conversion statuses of the clicks are also included in the data set.

In this data set, each row corresponds to a display ad chosen by Criteo and subsequently clicked by the user. The first two columns are click time and conversion time, where the latter is blank for unconverted clicks. The data set has 17 covariates (8 integer-valued and 9 categorical variables). Except for campaign ID (one of the categorical variables), the definitions of the covariates are undisclosed (due to confidentiality issues).

To evaluate the performance of our bias-adjusted estimators (i.e., E-bias-adjusted and W-bias-adjusted estimators) in this section, we investigate their bias, SE, and computation time relative to three other estimators: the \textbf{naive} estimator, the \textbf{oracle} estimator (the MLE of the logistic regression model based on the eventual conversion statuses of the clicks), and the maximizer of the DFM when the distribution of the delays is treated as exponential (Chapelle's estimator). Note that the oracle estimate is not obtainable in practice, where at any time $t$, the delay distribution parameters will be unknown. However, we include this estimator as a ``gold standard" to which we compare the other estimators.  

Following \citet{Chapelle}, we use log-loss to measure the bias of each estimator. Log-loss is a measure of the distance between parameter estimates and the true quantity of interest. In our case, log-loss is algebraically equivalent to the negative log-likelihood (NLL) of the logistic regression model (treating eventual conversion statuses as the true quantities of interest).

Estimating the parameters in the DFM can be very slow, depending on the number of covariates in the model. For instance, say we choose a subset of the covariates in the full data set such that we have 300 covariate coefficients (corresponding to the continuous covariates and the dummy variables that represent the categorical covariates) in the model. Obtaining the MLE of the DFM is approximately 500 times slower than computing the bias-adjusted estimator. 

To keep the parameter estimation time feasible in our data analysis, we first fit a logistic regression model with all possible covariates to the eventual conversion statuses of the clicks, using a LASSO penalty term with regularization parameter large enough that approximately 100 covariates appear  in the fitted model. We then use only the selected subset of the covariates in all of our analysis in this section. Note that the purpose of this variable selection is solely to facilitate estimation; in this paper, we are interested in the relative performances of the estimators given a set of covariates, not variable selection, per se. Therefore, we perform this variable selection only once.

Since the data set is huge, we use data splitting and use only a random sample (approximately $10\%$) of the data set as our training set. We then obtain our estimates on the training set and compute NLL on the rest of the data set (our test set). We repeat this procedure 40 times and report the average of the NLLs.

\begin{figure}
\centering
\includegraphics[width=10cm]{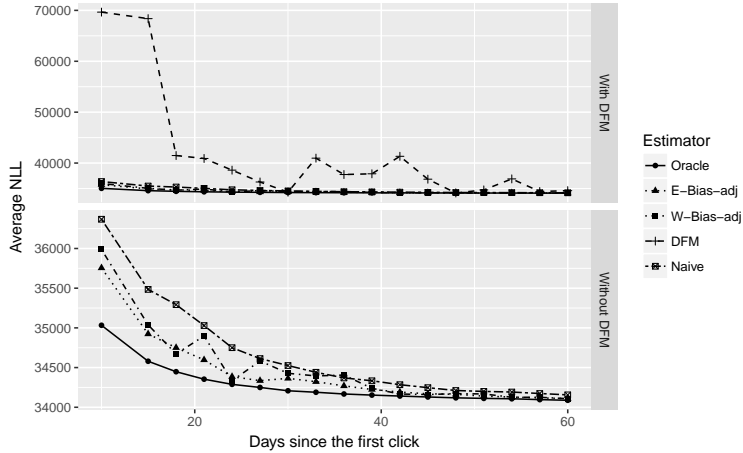}
\caption{Two views of the average NLL of the estimators at different time steps (averaged over 40 random splits of the data): with the DFM estimator (top) and without the DFM estimator (bottom)}
\label{NLLApp}
\end{figure}

Figure \ref{NLLApp} shows the average (over the 40 random splits of the data) NLL of the estimators at different time steps. The DFM estimator has convergence problems, especially when the number of known conversions is not large relative to the number of parameters (i.e., over the first two weeks of the observation period). After excluding the problematic estimates, the DFM estimator still behaves poorly (top plot of figure~\ref{NLLApp}). To illustrate the differences among the other estimators better, we omit the DFM estimator from the plot (bottom plot of figure~\ref{NLLApp}). The E-bias-adjusted estimator appears to outperform the other estimators. Specifically, the E-bias-adjusted estimator appears to outperform the W-bias-adjusted in the first month, and they perform similarly in the second month. In addition, as we obtain more new clicks and more information about the old clicks, the NLL of the estimators appears to decrease and get closer to that of the oracle estimator.

Table  \ref{TimeApp} shows the average  computation time (in seconds) of the estimates based on repeated data splitting when we have approximately 100 covariates in the model. The computation time of the DFM estimator is about 21 times longer than that of the E-bias-adjusted estimate, and the computation time of W-bias-adjusted estimator is about $30\%$ longer than that of the E-bias-adjusted estimator.
\begin{table*}
\caption{Average computation time (in seconds) of each estimate over repeated data splitting and different time steps. Approximately 100 covariates are included in the model.}
\label{TimeApp}
\centering
\begin{tabular}{l|r}
\hline
Estimator & Run time\\
\hline
$Naive$ & 4.67\\
$E-Bias-adjusted$ & 120.67\\
$W-Bias-adjusted$ & 158.30\\
$DFM$ & 2545.82
\\ \hline
\end{tabular}
\end{table*}

As a final note, the distribution of the observed delays of converted clicks looks closer to Weibull than exponential (see online material and also,  e.g., \citet{Ji:2016}) and thus the assumption of exponential delay times in the DFM is unreasonable. However, the MLE (i.e., of the maximizer of the model based on a Weibull distribution for the delays) has serious convergence issues and very long computation time. Thus we did not study the performance of this estimator in detail.
 
\section{Accuracy, precision and computational efficiency of the bias-adjusted estimators}
\label{Simulation}

In this section, we use a simulation study to evaluate the performance of our estimators. We investigate their bias, SE, and computation time. Besides the estimators mentioned in \S\ref{application}, since we know the delay distribution in our simulation study, we consider the \textbf{true-bias-adjusted} estimator (the bias-adjusted estimator computed using the true cdf of the delay distribution, so that the weights in (\ref{MultAdj}) are known). The true-bias-adjusted estimator helps us to gauge how much we lose by estimating the delay distribution parameters with \eqref{ExpAdj} (or \eqref{WeibAdj}).

We suggest using bias of the estimated probabilities as a measure of error in the simulation study. Average bias at time $t$ is defined as $\frac{1}{n} \sum_i\left(p_i - \hat{p}_i\right)$ for an estimator of $p_i$, $\hat{p}_i$. Recall that the $p_i$'s vary according to covariates; average bias can be interpreted as an estimate of the {\it marginal bias} of the estimator of probability of conversion (in contrast with ${\rm E}[p_i-\hat{p}_i]$, which represents the bias of $\hat{p}_i$ {\it conditional} on ${\bm x}_i$).

\subsection{Simulation study design}

Since our focus in this paper is display advertising, we use a real data set (the Criteo data described in Section~\ref{application}) to inform the design of our simulation study. Specifically, we pick approximately 8500 clicks ($n \approx 8500$) from a campaign with a large number of clicks. For this campaign, the average conversion probability was moderate ($\sim 30 \%$). We use the covariates values given in the Criteo data set by \citet{Chapelle} and keep these values the same across runs. Since for the selected clicks some covariates have only one value (or have only a few values that differ from the mode), we use only three of the categorical variables (resulting in 16 dummy variables) and four of the integer-valued covariates in the original data set.

We conduct two simulation studies. In the first, we generate exponential-distributed conversion delays. In the second, we generate Weibull-distributed conversion delays. The parameters of these distributions are set to their estimated values based on the observed delays in the chosen campaign (using only converted clicks). In other words, the estimated parameters based on the Criteo data become the true parameter values in the simulation study. Similarly, we estimate the regression coefficients of the conversion probability model by fitting a logistic regression to the final conversion status of the clicks in the data set. We then use these estimated coefficients as the true coefficients in the simulation studies (see online material for the covariates and coefficients we use).

We consider two factors affecting the performance of the conversion probability estimators: average conversion probability and average delay time, where average means across all clicks of the campaign. We choose the levels of the factors based on the range of the conversion probabilities and delays in the real data set; see table \ref{Factor} for details. To keep the simulation study feasible and the number of parameters in the model manageable, we assume no interaction among the covariates -- in particular no interaction between campaign and the other covariates. Under this assumption, we can vary the factors of interest (average conversion probability and average delay time) simply by varying the values of the intercepts and of the campaign effects in both the delay and conversion models while keeping the other covariate coefficients (and the shape parameter, in the Weibull case) fixed.
\begin{table*}
\centering
\begin{threeparttable}
\caption{Levels of the factors in the simulation studies}
\label{Factor}
\begin{tabular}{l|ccc}
\hline
Factor & Low & Medium & High\\
\hline
Conversion probability\tnote{\textdagger} & 0.1 & 0.3 & 0.6\\
Delay mean\tnote{*} & 2 & 4 & 7
\\ \hline
\end{tabular}
\begin{tablenotes}
\item[\textdagger] averaged across all clicks
\item[*] in days
\end{tablenotes}
\end{threeparttable}
\end{table*}

To create a realistic scenario in our simulation studies, we track the clicks since start of the data collection at $t = 0$, and evaluate the estimators at 17 different time steps over a two month period (with time steps spaced far enough apart such that approximately equal numbers of clicks occur in each interval). At each time step $t$, we consider only clicks that occurred by $t$. Similarly, we treat a click as converted only if we observe its conversion by $t$ and its age is less than $W = 30$ days. Otherwise, we treat it as unconverted.

\subsection{Study 1}

We first consider the case where the conversion delays follow an exponential distribution. In other words, we generate data from Chapelle's DFM. Thus, the MLE of the DFM and the E-bias-adjusted estimator are based on the correct model.

\begin{figure}
\centering
\includegraphics[width=10cm]{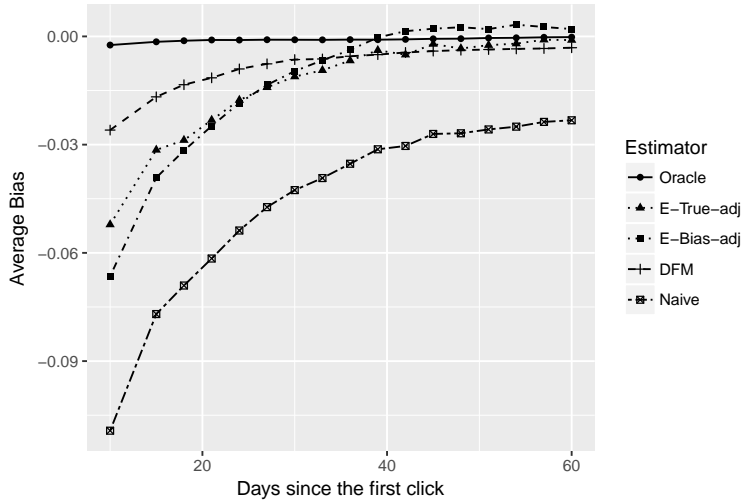}
\caption{Average bias of the estimators over time for the medium level of the factors when the delays follow an exponential distribution}
\label{BiasExp}
\end{figure}

Figure \ref{BiasExp} shows the average bias of the estimators over time when both factors (average conversion probability and average delay) are at their medium level. As expected, since the DFM is the true model in this study, its maximizer (the MLE) outperforms all other estimators (except the oracle estimator). In particular, it appears to be less biased than the E-bias-adjusted estimator (especially over the first month). That said, the bias of both estimators seems quite small in the second month (less than 0.007 on average). The true-E-bias-adjusted estimator appears to perform slightly better than the E-bias-adjusted estimator (especially over the first month), and the naive estimator appears to remain biased even after two months by approximately 0.05. The overall trend in bias is similar when we use other levels of the factors given in table \ref{Factor}. As expected, when the average delay is at its low level, the accuracy of the naive estimator appears to be almost as high as the other estimators. Moreover, the MLE of the DFM behaves poorly when the average delay is high and average conversion probability is low (see online materials).

\subsection{Study 2}

In this study, we consider a Weibull distribution for the delays.

We compute all the estimators (including the W-bias-adjusted estimator) over time as in study~1. Note that in this case, the maximizer of the DFM and the E-bias-adjusted estimator are both based on the (misspecified) exponential distribution for the delay times. Thus, the former is no longer the MLE and we call it DFM estimator in this study. We do not study the MLE (i.e., the maximizer of the DFM modified to allow a Weibull distribution for the delays) due to convergence issues and very long computation times. In addition, Since the true-E-bias-adjusted estimator hasn't been derived for this study, we don't consider the estimator here.

\begin{figure}
\centering
\includegraphics[width=10cm]{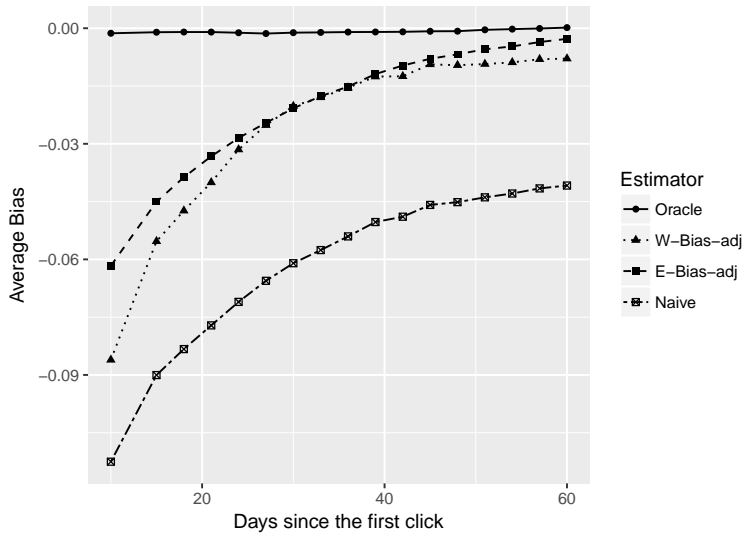}
\caption{Average bias of the bias-adjusted estimators over time for the medium level of the factors when the delays follow a Weibull distribution}
\label{WeibExpBias}
\end{figure}

Figure \ref{WeibExpBias} shows the average bias of the E-bias-adjusted and W-bias-adjusted estimators, along with that of the oracle and naive estimators over time when both factors (average conversion probability and average delay) are at their medium level. Over the first three weeks, the E-bias-adjusted estimator appears to slightly outperform the W-bias-adjusted estimator. However, both estimators perform similarly after the third week. In addition, the computation time of the W-bias-adjusted is approximately $30\%$ more than that of the E-bias-adjusted estimator. Therefore, we consider only the E-bias-adjusted estimator for the remainder of this paper.

\begin{figure}
\centering
\includegraphics[width=10cm]{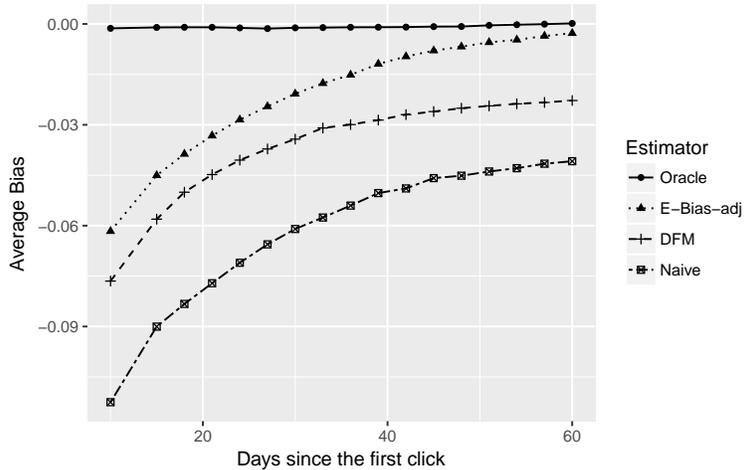}
%\figurebox{18pc}{30pc}{}[MAE.eps]
\caption{Average bias of the estimators over time for the medium level of the factors when the delays follow a Weibull distribution}
\label{WeibBias}
\end{figure}

Figure \ref{WeibBias} shows the bias of the estimators over time when both factors, average conversion probability and average delay, are at their medium level. In contrast with study~1, the E-bias-adjusted estimator appears to outperform the DFM estimator. In particular, as time goes on, the bias of the E-bias-adjusted estimator nearly disappears, whereas the bias of the DFM estimator does not. In addition, the bias of the E-bias-adjusted estimator shows that the maximum penalized likelihood estimator of the parameters in the delay time model (see \eqref{ExpAdj}) performs well even when the delay distribution is misspecified, especially for $t \geq 30$. The trend in bias is similar for other levels of the factors given in table \ref{Factor}. Again, when the delay mean is in its low level, the accuracy of the naive estimator is almost as high as the other estimators. Similar to study~1, the DFM estimator behaves poorly when the average delay is high and average conversion probability is low (see online materials).

\subsection{Coverage probability of the bias-adjusted estimator}

As mentioned in \S\ref{BAestimator}, we can efficiently compute a SE for $\hat{p}_i$ as a function of the derivative of the left side of \eqref{fakescore}. In this section, we study the validity of our SE.

%Since the equations in \eqref{MultAdj} are algebraically equivalent to the weighted quasi-score equations, we can efficiently obtain a standard error (SE) for the E-bias-adjusted estimator. In addition, because our estimating equations \eqref{MultAdj} are asymptotically equivalent to the score equations associated with a logistic regression model, under mild conditions, we can derive a consistent estimator of the standard deviation of $\hat{\beta_c}$.

\begin{figure}
\centering
\includegraphics[width=10cm]{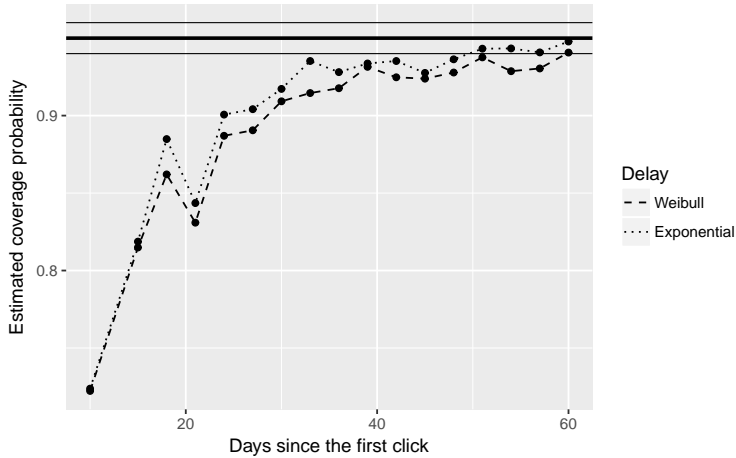}
\caption{Average coverage probability of the $95\%$ CI for conversion probability based on the E-bias-adjusted estimator over time for the medium level of the factors (studies~1 and~2)}
\label{WeibCP}
\end{figure}

Figure \ref{WeibCP} shows the average coverage probability (CP) associated with 95\% confidence intervals for conversion probability based on the E-bias-adjusted estimator over time when the delays follow  exponential or Weibull distributions. In the first month, the average CP is below the nominal level (approximately $88\%$). However, in the second month, the average CP is more than $92\%$. To show the closeness of the average CP to the nominal value of $0.95$ at each time point more carefully, we add the non-rejection region for the score test of whether CP differs from 0.95.  This region is defined as $(0.95 - 2\sqrt{0.95(1 - 0.95) / R} , 0.95 + 2\sqrt{0.95(1 - 0.95) / R}) \approx (0.94 , 0.96)$, where $R = 2000$ is the number of replicates. The CP when the delays are exponential-distributed (so that the E-bias-adjusted estimator is based on the correct model) is not significantly different the nominal coverage level at the last 4 time steps. In contrast, when the delays are Weibull-distributed,  the CP differs significantly from $0.95$ except at the last time step. In other words, CP is lower when the E-bias-adjusted estimator is based on a misspecified model, but our results suggests that it converges to $0.95$.

To compute the SE (and CP) associated with the DFM estimator, we could compute the Hessian matrix of the estimates for each replicate. However, this matrix was non-positive-definite for most replicates. For this reason, we omit results concerning the DFM estimator. The results for the other runs are similar (see online material).

\subsection{Computation time}

Given the very short time available for choosing an ad and publishing it on the host website -- and the huge number of ad requests and new campaigns at any time -- publishers need to refit the model and obtain the conversion probabilities frequently. Therefore, computation time is a critical issue in display advertising. Table  \ref{Time} shows the average computation times of the estimates along with their sample standard deviation (SSD) when the true delay distribution is Weibull (Study~2) and the factors are at their medium level. In particular, the computation time of the DFM estimator is more than 5 times that of the E-bias-adjusted estimate. For the levels of the factors that we considered, this ratio can be between 4 and 8.
\begin{table*}
\caption{Average computation times (in seconds) of each estimator over different time steps along with its SSD for the medium level of the factors (Study~2)}
\label{Time}
\centering
\begin{tabular}{c|c}
\hline
Estimator & Run time average (SSD)\\
\hline
$Naive$ & 0.09 (0.04)\\
$Bias-adjusted$ & 5.16 (2.38)\\
$DFM$ & 24.54 (4.02)
\\ \hline
\end{tabular}
\end{table*}
The computation time of the estimates is similar for Study~1.

\section{Discussion}
\label{Discussion}

In this paper, we developed a method for estimating probability of conversion efficiently and with high accuracy. In particular, we introduced a bias-adjusted estimator based on a simple (misspecified) logistic model, and evaluated its accuracy and computational efficiency.

As an alternative, we could obtain the MLE and bias-adjusted estimators by assuming a Weibull distribution for the delays, which would allow greater flexibility in the model and would, in particular, provide a better description of the the delays in the Criteo data (see online material). However, the MLE of this model suffers from both convergence issues and lengthy computation times. Moreover, the W-bias-adjusted estimator is not as consistent and efficient as the E-bias-adjusted estimator. Therefore, we recommend the E-bias-adjusted estimator even when the delays follow a Weibull distribution.

Since clicks have different associated true probability of conversion, the estimators of these probabilities (and their bias) have different variances. When computing the average of bias, one may account for these differences by weighting each bias by its true SD, especially when the range of the true probabilities is large. In our case, there was no difference between behaviour of the estimators in bias and weighted bias.

To reduce overall computation time in the example, we used data splitting to obtain the estimates in our application. Comparing the performance of the estimators over the entire data set could be another interesting problem.

Our estimation method incorporates data only from users' final click on an ad. In other words, we ignore users' previous (unconverted) clicks on the same ad. Interesting future work could be a model that can capture the information in the historical unconverted clicks of the users.

\appendix

\section{Likelihood of Z}\label{appendix}

To prove \eqref{TrueModel}, we first derive the cdf of $Z_i$ as
\begin{eqnarray}
G_{Z_{i}} (z_{i}) &=& P \left( Z_{i} \leq z_{i} \right) \\ \nonumber
&=& P \left( Z_{i} \leq z_{i} | C_{i} = 1 \right) P(C_{i} = 1) \\ \nonumber
&+& P \left( Z_{i} \leq z_{i} | C_{i} = 0 \right) P(C_{i} = 0) \\ \nonumber
&=& \left\{
	\begin{array}{ll}
		0  & \mbox{if } z_{i} \leq 0 \\
		H_i (z_{i}) p_i  & \mbox{if }  0 < z_{i} < a_{i} \\
		1 & \mbox{if } z_{i} \geq a_{i}
	\end{array}
\right. ,
\end{eqnarray}
where $p_i = \frac{\exp(\bm{\beta_c}^\prime \bm{x_i})}{1 + \exp(\bm{\beta_c}^\prime \bm{x_i})}$. Therefore, the likelihood function is
\begin{eqnarray}
L_g \left(\bm{\beta_c} | \bm{z} \right) &=& g \left( \bm{z} \right) \\ \nonumber
&=& \prod_{i} g \left( z_{i} | a_{i} \right) \\ \nonumber
&=& \prod_{i} \left( p_i h(z_{i}) \right) ^ {I( z_{i} < a_{i})} \left( 1 - p_i H(z_{i}) \right) ^ {I( z_{i} \geq a_{i})} .
\end{eqnarray}

\end{document}